%% file: conference_101719.tex
\documentclass[conference]{IEEEtran}
\IEEEoverridecommandlockouts
\usepackage{cite}
\usepackage{amsmath,amssymb,amsfonts}
\usepackage{algorithmic}
\usepackage{graphicx}
\usepackage{hyperref}
\usepackage{textcomp}
\usepackage{xcolor}
\usepackage{xspace}
\usepackage{lipsum}
\usepackage{subcaption} 
\usepackage{array} 
\usepackage{multirow}
\newtheorem{definition}{Definition}
\newcommand{\methodname}{Geo-Llama\xspace}
\newcommand{\partitle}[1]{
  \smallskip
  \noindent
  \textbf{#1}}
\usepackage{balance}
\usepackage{todonotes}
\usepackage{newfloat}
\usepackage{listings}
\usepackage{comment}
\usepackage{url}
\usepackage{makecell}

\def\BibTeX{{\rm B\kern-.05em{\sc i\kern-.025em b}\kern-.08em
    T\kern-.1667em\lower.7ex\hbox{E}\kern-.125emX}}
\begin{document}

\title{Geo-Llama: Leveraging LLMs for Human Mobility Trajectory Generation with 
Constraints\\
}

\author{
\IEEEauthorblockN{Siyu Li\IEEEauthorrefmark{1}, Toan Tran\IEEEauthorrefmark{1}, Haowen Lin\IEEEauthorrefmark{2}, John Krumm\IEEEauthorrefmark{2}, Cyrus Shahabi\IEEEauthorrefmark{2},}
\IEEEauthorblockN{Lingyi Zhao\IEEEauthorrefmark{3}, Khurram Shafique\IEEEauthorrefmark{3}, Li Xiong\IEEEauthorrefmark{1}}
\IEEEauthorblockA{\IEEEauthorrefmark{1}Dept. of Computer Science, Emory University, Atlanta, GA, USA}
\IEEEauthorblockA{\IEEEauthorrefmark{2}Dept. of Computer Science, University of Southern California, Los Angeles, CA, USA}
\IEEEauthorblockA{\IEEEauthorrefmark{3}Novateur Research Solutions, Ashburn, VA, USA}
\IEEEauthorblockA{
\{siyu.li, viet.toan.tran, lxiong\}@emory.edu, \{haowenli, jckrumm, shahabi\}@usc.edu, \{lzhao, kshafique\}@novateur.ai
}
}

\maketitle

\begin{abstract}
Generating realistic human mobility data is essential for various application domains, including transportation, urban planning, and epidemic control, as real data is often inaccessible to researchers due to high costs and privacy concerns. Existing deep generative models learn from real trajectories to generate synthetic ones. Despite the progress, most of them suffer from training stability issues and scale poorly with increasing data size. More importantly, they often lack control mechanisms to guide the generated trajectories under constraints such as enforcing specific visits. 
To address these limitations, we formally define the controlled trajectory generation problem for effectively handling multiple spatiotemporal constraints. We introduce Geo-Llama, a novel LLM finetuning framework that can enforce multiple explicit visit constraints while maintaining contextual coherence of the generated trajectories. 
In this approach, pre-trained LLMs are fine-tuned on trajectory data with a visit-wise permutation strategy where each visit corresponds to a specific time and location. This strategy enables the model to capture spatiotemporal patterns regardless of visit orders while maintaining flexible and in-context constraint integration through prompts during generation. 
Extensive experiments on real-world and synthetic datasets validate the effectiveness of Geo-Llama, demonstrating its versatility and robustness in handling a broad range of constraints to generate more realistic trajectories compared to existing methods. 
\end{abstract}


\input{1.introduction}
\input{2.relatedwork}
\input{3.method}
\input{4.experiment}

\balance
\input{5.conclusion}
\input{acknowledgment}

\newpage
\bibliographystyle{IEEEtran}

\balance
\bibliography{references}

\end{document}

%% file: 1.introduction.tex
\begin{figure*}[ht]
    \centering
    \includegraphics[width=1.0\linewidth]{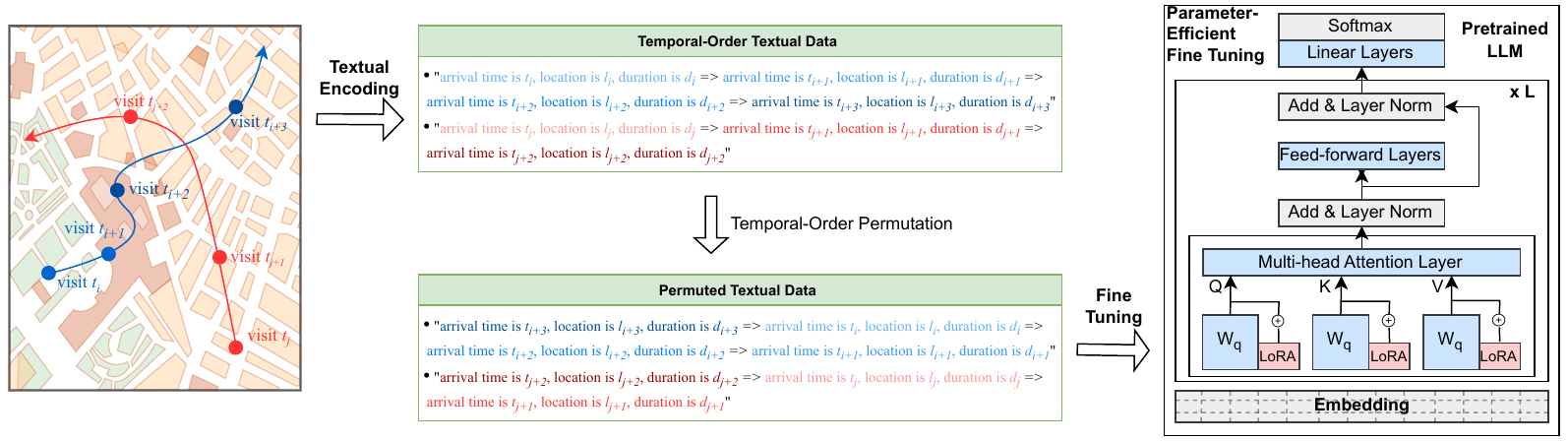}
    \vspace{-3mm}    
    \caption{Fine tuning mechanism of \methodname. First, the trajectories are converted into text strings through Textual Encoding. Next, the Temporal-Order Permutation permutes the textual data. The permuted data is then used for fine tuning.}
    \label{fig:method-finetuning}
    \vspace{-2mm}
\end{figure*}

\begin{figure*}[ht]
    \centering
    \includegraphics[width=1.0\linewidth]{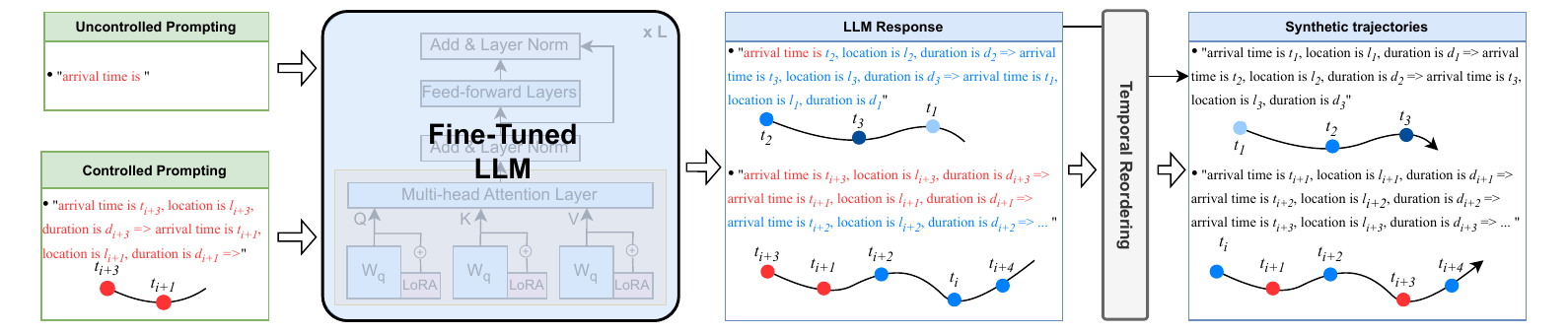}
    \vspace{-3mm}        
    \caption{The generation mechanism of  \methodname. It supports both uncontrolled and controlled generation. It generates responses based on the input prompts which are reordered based on the temporal order of visits' arrival time to achieve the final trajectories.}
    \label{fig:sampling}
    \vspace{-5mm}
\end{figure*}

\section{Introduction}

Human mobility data, represented as trajectories or sequences of visits, is essential for advancing research and applications in transportation, urban planning, social dynamics, and epidemiology \cite{hu2021human, becker2011tale, chen2019human}. However, the collection of large-scale trajectory data is often limited by high costs and privacy concerns, which makes it inaccessible to researchers \cite{10.1145/3652158}. This gives rise to an important research area of generating synthetic but realistic trajectories.

One approach for synthetic trajectory generation involves developing micro-simulators, which require careful parameter calibration on sensor data, traffic patterns, and crowd movement statistics \cite{LE201681}. The resulting parameters often lack accuracy and are challenging to configure manually, as such traditional simulation methods relying on heuristics fail to capture complex human mobility patterns \cite{feng2020learning}. 
Data-driven approaches such as generative adversarial networks (GANs) and variational point process can generate large-scale and realistic trajectories by learning from real-world data \cite{lin2023generating,zhang2023csgan,ouyang2018non,long2023practical}. However, many existing methods struggle with training stability and do not scale well as data size increases, resulting in the loss of fine-grained spatiotemporal patterns. 

Another significant limitation of existing techniques is their lack of control over the trajectory generation process. Incorporating control in trajectory generation to enforce specific visits is crucial as it enables various scenario analysis for downstream tasks. 
For example, to use synthetic mobility data for epidemiology studies, we may need to generate 
trajectories that pass through a school building during certain times of the day to simulate school being open and trajectories staying home to represent school closure. This enables analysis of potential outcomes of alternative interventions, which is essential for informed decision-making  in fields such as public health. Moreover, by synthesizing trajectories that adhere to specific patterns based on prior knowledge that might be available, the generated data aligns more closely with both real-world behaviors and individual preferences.
 

Incorporating spatiotemporal constraints such as specific visits in trajectory generation is particularly challenging and remains largely unexplored. 
 Recent studies on controllable trajectory generation have focused on integrating structural constraints, such as road network, or ensuring moving validity by avoiding overlaps with obstacles \cite{zhu2024controltraj, zhang2020factorized}. These are confined to implicit constraints, typically at a global or route level, and therefore cannot be generalized to enforce explicit visit-level constraints. 


A recent work leverages transformer-based models to complete trajectories by infilling missing visits among given visit constraints \cite{hsu2024trajgpt}. While it represents a significant step towards explicitly incorporating visit constraints into mobility data generation, it is only effective in handling a large number of given constraints (known visits) due to its imputation nature and does not generate realistic trajectories when there are only a few visit constraints. Another recent work attempted to propagate a visit constraint into time and location constraints for all other visits, 
allowing constrained time and location sampling for each visit \cite{lin2024controllable}. The complexity of such propagation restricts this approach to handling only a single constraint, limiting its applicability in more complex scenarios.
Alternatively, rather than incorporating control directly during the generation stage, a potential approach \cite{zhang2023survey} is to train the generator without explicit constraints and then apply post-processing to select samples with desired constraints from a large pool of uncontrolled outputs. However, it is highly unlikely for an unconstrained trajectory from a large combinatorial space to satisfy a very specific visit constraint.

\smallskip
\noindent
\textbf{Contributions}. In this paper, we study the problem of controlled generation of human mobility trajectories with multiple visit constraints. Our trajectories are specified by a series of visits, where each visit is defined by a location and arrival time. 
Our constraints include one or more required visits with specific time and location ranges.
The controllable generation should ensure that both the visit constraints and the statistical properties of the training data are maintained. This will help preserve context-coherent transitions between the specified visits and other visits.

To solve this problem, we propose Geo-Llama, a framework for generating human mobility data with spatiotemporal constraints leveraging large language models (LLMs). The main idea is to represent trajectories as sequences of tokens and fine-tune LLMs on next-token prediction tasks. This process captures the inherent spatiotemporal correlations and movement patterns (as shown in Figure \ref{fig:method-finetuning}). The fine-tuned model can then be used 
to generate context-coherent trajectories to support both uncontrolled generation and controlled generation that satisfies specified constraints (Figure \ref{fig:sampling}). 
\\
\indent
Specifically, to enable the model to capture true movement patterns based on time features (regardless of the order of the visits in the sequence), we propose a visit-wise permutation strategy that randomly permutes the visits in a sequence 
before using them for fine-tuning the LLMs. In other words, we can consider the problem as learning the spatiotemporal correlations from a set of visits with time and location instead of an ordered sequence of visits.  This also enables us to generate a trajectory in which the constrained visit may occur anywhere in the sequence. 
The generated sequences can be simply reordered based on the time features into ordered visits.  Lastly, a thoroughly designed integrity-check algorithm is employed during post-processing to ensure the consistency and coherence of the trajectories (e.g., by preventing overlapping time intervals between consecutive visits). 
\\
\indent Our major contributions can be summarized as follows:
\begin{itemize}

    \item We formulate the controlled trajectory generation problem with multiple visit constraints. The goal is to generate realistic trajectories that represent the patterns in the training data while enforcing the specific visit constraints. 
    \item We introduce Geo-Llama, the first general LLM-based fine-tuning framework for synthetic trajectory generation. It supports both uncontrolled and controlled generation. 
    It utilizes a visit-wise permutation strategy for encoding the trajectories and leverages LLMs to capture the inherently complex spatiotemporal patterns underlying human mobility trajectories while enforcing multiple  constraints in a context-coherent manner.
    \item We conduct extensive experiments comparing \methodname with state-of-the-art generation methods on real-world and synthetic datasets. We demonstrate that \methodname not only enables controlled synthetic trajectory generation but also generates the most realistic trajectories in both controlled and uncontrolled settings. Moreover, \methodname achieves higher data efficiency compared to existing approaches.
    
\end{itemize}

\vspace{-2mm}








%% file: 2.relatedwork.tex
\section{Related Work}


\noindent
\textbf{Human Mobility Trajectory Generation.} Conventional simulation approaches generate human mobility trajectories using physical estimations based on parameters collected by sensors and surveys \cite{pelekis2013hermoupolis,jiang2016timegeo}. These methods heavily depend on human heuristics and are error-prone. Recent works leveraged large language models (LLMs) for agent based modeling and simulation but simulating large-scale LLM agents remains highly challenging \cite{gao2024large}.

Deep learning and generative models directly learn and sample trajectories from data-driven latent representations. Early works used recurrent neural networks (RNNs) to predict (generate) the next visited location by modeling spatiotemporal transitions \cite{liu2016predicting,feng2018deepmove}. These models primarily focus on short-term predictions and struggle to generate long, coherent trajectories that capture real-world patterns over time. Generative adversarial networks (GAN) \cite{goodfellow2020generative} based approaches are becoming prevalent owing to their strong distribution modeling capability via a mini-max game between the generator and discriminator. For example, TrajGAN \cite{ouyang2018non} converts trajectories into two-dimensional images and uses CNN-based GANs for generation. SeqGAN \cite{yu2017seqgan} proposes a reinforcement learning framework that adopts the discriminator outputs as rewards augmented by the Monte-Carlo rollout to provide fine-grained supervision on the generator's next location prediction. MoveSim \cite{feng2020learning} extends SeqGAN by leveraging self-attention networks as the generator backbone and incorporating prior knowledge of human mobility patterns such as Points of Interest (POIs) into the generation process. CSGAN \cite{zhang2023csgan} addresses the sampling diversity issues by clustering different motion modalities, enabling more granular learning of mobility patterns and enhancing the realism of generated trajectories. Despite the progress, these approaches still struggle with unstable training and poor scalability as data size increases. Their  temporal encoding with fixed time intervals (e.g., every 15 minutes) results in computationally intensive and unnecessarily lengthy sequences, reducing representation efficiency. Such long sequences with repeated locations (corresponding to a stay point) also make it challenging to enforce specific visit-level constraints. To address such limitations, recent works adopt the variants of Deep Spatiotemporal Point Process (DeepSTPP)\cite{zhou2022neural} for trajectory generation \cite{long2023practical,zhou2022neural}. Such approaches represent the trajectory as a sequence of distinct visit points instead of the fixed-interval representation, and therefore better capture the moving patterns by jointly modeling spatial and temporal factors. All of these methods do not consider constraints, and we will compare our approach with representative ones under uncontrolled setting. 

Recently, LLMs have demonstrated powerful sequence generation capabilities, enabling the understanding of complex contexts and the utilization of rich prior knowledge. A recent work \cite{bhandari2024urban} applies prompt engineering on pre-trained LLMs for trajectory generation by crafting targeted prompts using statistical summaries of demographic information, event types, and event-temporal correlations, emulating travel diaries without requiring additional model training. Another work \cite{wang2024large} presents a self-consistent activity pattern identification framework, where the LLM leverages its prior knowledge to evaluate all potential user profiles and POI background information paired with a specific trajectory, resulting in optimal prompt combinations. Retrieval-augmented generation (RAG) is then used to refine trajectory generation within the prompt-engineering paradigm. While these methods rely on rich contextual data like user attributes and POI details, their prompt-engineering nature indicates that the geographic coordinates or temporal dependencies are not explicitly modeled, limiting their applicability for datasets without augmented info. In contrast, our approach directly learns from timestamps and coordinates with LLM fine-tuning, enabling fine-grained modeling of spatiotemporal patterns without requiring extensive external information. Therefore, we do not consider these methods as baselines for comparison.













\vspace{0.7mm}
\noindent
\textbf{GPS Trajectory Generation.} 
While human mobility trajectory generation focuses on discrete movement events (usually corresponding to a number of stay points per day) and their spatiotemporal relationships, GPS trajectory generation focuses on producing fine-grained sequences of geographic coordinates at regular, short intervals (e.g., every second). 
Early approaches \cite{armstrong1999geographically, zandbergen2014ensuring} relied on perturbing real trajectories with noise or combining trajectory segments from real data to generate synthetic trajectories, which often alter spatial-temporal characteristics and compromise data utility.
GAN-based methods \cite{wang2021large, cao2021generating} are also proposed for GPS trajectory generation and suffer similar issues as for human mobility generation. 
DiffTraj \cite{zhu2023difftraj} introduces a spatial-temporal diffusion probabilistic model combined with a Trajectory-specific UNet to effectively capture spatiotemporal patterns from GPS trajectories, ensuring stable training while preserving the original geographic distribution and supporting downstream tasks.
Given the distinct objectives and granularity, we do not include GPS trajectory generation methods in our comparisons.

\noindent
\textbf{Controlled Mobility Trajectory Generation}.
In controlled settings, some works start incorporating constraints tailored to domain-specific requirements. FDSVAE \cite{zhang2020factorized} learns and operates on latent representations to enforce spatiotemporal validity constraints. ControlTraj \cite{zhu2024controltraj} is a diffusion-based trajectory generation framework that enforces road network topology constraints to adapt to new geographical contexts. Another line of work \cite{takagi2024hrnet,wang2023privtrace} addresses privacy concerns in synthetic trajectory generation by enforcing differential privacy constraint, focusing on improving privacy and accuracy tradeoff. These constraints are either implicit or at a global or route level instead of at a visit level.  

To accommodate explicit visit-level constraints, Traj-GPT \cite{hsu2024trajgpt} treats spatiotemporal constraints as fixed templates and leverages encoder-decoder transformer-based models to learn the spatiotemporal statistics of trajectory to fill the gaps. Due to the infilling nature, such an approach is more effective in handling dense constraints (e.g., their experiments masked 20\% visits for infilling while having 80\% as constraints) but is not suitable when constraints are sparse  or in uncontrolled setting when there is no constraint. 
Therefore, we will not compare it with our approach as a baseline. 
To incorporate explicit spatiotemporal constraints in a generative setting, Geo-CETRA \cite{lin2024controllable} introduces constraint factorization into spatiotemporal point processes, 
allowing in-context adjustments to all visits conditioned on the constraint visit. However, the inflexibility of factorization restricts this approach to handling only a single constraint. We will use this as a baseline for comparison under one constraint but our approach is more flexible and can handle multiple constraints. 

\vspace{0.7mm}
\noindent
\textbf{Controlled Sequence Generation in Other Domains.} 
Beyond mobility data, there are controllable generation work focusing on image data, using constraints like labels or properties, but these methods are not suitable for sequence data with specific constraints \cite{wang2022controllable}. 
One potentially related line of work is in the natural language processing (NLP) area which aims to generate text that adheres to 
specific linguistic and structural constraints for downstream applications \cite{zhang2023survey}. Most of these approaches only handle semantic-level constraints which are too implicit to accommodate specific hard requirements such as placing a specific word in a precise position within the text. Some approaches \cite{donahue2020enabling} formulate hard constraint enforcement as an infilling problem, but rely on fixed positions, making them not readily applicable for mobility generation with visit constraints. 

\vspace{0.7mm}
\noindent
\textbf{Next Point-of-Interest (POI) Recommendation.} 
Related to the task of generation, next POI recommendation predicts a user’s next point of interest (POI) based on historical mobility data and contextual information, often framed as a sequence recommendation task. A variety of methods have been proposed including probabilistic models, deep learning models, graph-based models, and LLMs\cite{cheng2013you, kong2018hst, sun2020go,luo2021stan, zhang2022next,lim2020stp, yang2022getnext, zhang2022next,li2024large}. 
Unlike the next POI recommendation, human mobility trajectory generation focuses on producing entire trajectories, often from noise, without requiring historical inputs. 
Although the sequence-based POI recommendation models can be extended autoregressively to construct full trajectories, they often fail to capture global dependencies. Given these fundamental differences, we do not consider next POI recommendation models as baselines for our work.

%% file: 3.method.tex
\section{Preliminaries}
In this section, we define the terminology and our problem formulation. 

\begin{definition}
  A \textbf{\textit{Trajectory Dataset}} is a set of trajectories denoted as $\mathbb{D} = \{\tau^{(i)}| i = 1,..,N\}$, where $\tau^{(i)}$ stands for the $i$-th trajectory and $N$ is the number of trajectories.
\end{definition}

\begin{definition}
    A \textbf{\textit{Trajectory}} is defined as a sequence of visits through space and time, denoted as $\tau = \{v_j \mid j = 1, \dots, L\}$, where each visit $v_j = (t_j, l_j, d_j)$ consists of the arrival time $t_j \in \mathbb{R}$, the location $l_j \in \mathcal{L}$, and a feature vector $d_j \in \mathbb{R}^p$ representing attributes such as stay duration. Here, $L$ is the total number of visits in trajectory $\tau$.
    
    In addition to the visit-based representation, a trajectory can also be expressed in a \textbf{\textit{fixed-time-interval representation}} as $\tau' = \{l_k \mid k = 1, \dots, T\}$, where $T$ is the total number of time intervals, and $l_k \in \mathcal{L}$ is the location associated with the $k$-th time interval. 
    For example, if we model daily trajectories, and use fixed time interval $\Delta t = 15$ minutes, it results in  $T = \frac{24 \times 60}{\Delta t} = 96$. Consider a trajectory in the visit-based representation $\tau = \{..., (10\text{am}, l_1), (11\text{am}, l_2), ...\}$, the corresponding fixed-time-interval representation is $\tau' = \{..., l_1, l_1, l_1, l_1, l_2, l_2, l_2, l_2, ...\}$. 
    


    Fixed time interval representation creates redundant lengthy sequences and inflexibility in constraint enforcement. Visit-based trajectories offer flexible, compact representations, which we adopt for this work.
\end{definition}

\begin{definition}
A \textbf{\textit{Spatiotemporal Visit Constraint}} is a required visit  denoted as $c = (l_c, t_{c1}, t_{c2})$, where $l_c$ is a specific location, [$t_{c1}$, $t_{c2}$] is the range for the arrival time. A trajectory $\tau$ satisfies a visit constraint $c$ if: $\exists\; v_i \quad \text{s.t.} \quad l_i = l_c \quad \& \quad t_{c1} \leq t_i \leq t_{c2}$.
i.e. the trajectory must contain a visit to $l_c$ that arrives between $t_{c1}$ and $t_{c2}$.
\label{def:constraint}
\end{definition}


\begin{definition}
\textbf{\textit{Controlled Synthetic Trajectory Generation}} is to generate a synthetic set $\mathbb{D}'$ which preserves the spatiotemporal patterns of the original dataset $\mathbb{D}$ while satisfying a set of given visit constraints. The problem can be formalized as Equation~\ref{eq:prob}, where $\psi$ is a function that measures the similarity between the trajectory properties of $\mathbb{D}'$ and $\mathbb{D}$, such as distributional divergence (e.g., JSD) over temporal, spatial, or structural statistics. While in rare cases where the constraint set $C$ happens to exactly match a subset of $\mathbb{D}$, the optimal solution $\mathbb{D}'$ may coincide with $\mathbb{D}$, the more typical scenario involves partial or contextually shifted constraints, requiring the model to generate trajectories that resemble but are not identical to the original dataset.

\vspace{-3mm}

\begin{equation}
\small
    \begin{aligned}
    \textup{maximize} \quad & \psi(\mathbb{D}', \mathbb{D}) 
    \quad \textup{s.t.} \quad & \mathbb{D}' \; \textup{satisfies} \; C
    \end{aligned}
    \label{eq:prob}
    \vspace{-2mm}
\end{equation}

\end{definition}

\begin{definition}
  \textbf{\textit{Time \& Location Discretization}}: The location domain is divided into equal-sized grid cells, each represented by a unique ID from the set $\mathcal{G} = \{g_1, \dots, g_{|\mathcal{G}|}\}$. Time domain is discretized into fixed 
  intervals, forming the set $\mathcal{I} = \{i_1, \dots, i_{|\mathcal{I}|}\}$. After discretization, each location $l_j \in \mathcal{G}$ and time point $t_j \in \mathcal{I}$ represent distinct tokens in our trajectory dataset, reducing complexity and leveraging the model's strength in handling categorical data.
  \end{definition}

\section{Methodology}

In this section, we describe our \methodname framework which leverages pretrained LLMs for realistic synthetic trajectory generation. There are two main stages: fine-tuning and generation.

\subsection{Fine-Tuning} Figure~\ref{fig:method-finetuning} illustrates the fine-tuning mechanism of \methodname. Each trajectory is first converted into textual data. The temporal order of the sequences is then permuted to enable the LLM to learn the spatiotemporal correlations rather than the order of the visits. A pre-trained LLM is fine-tuned using parameter-efficient techniques. This study focuses on decoder-only LLMs, which represent the state-of-the-art architecture for many generation tasks~\cite{llmsurvey}.

\partitle{Textual Encoding}. Textual Encoding converts trajectory $\tau^{(i)}$ into textual representation $\mathbb{S}^{(i)}$. To simplify, we consider $d_j$ of visit $(t_j, l_j, d_j)$ as a univariate feature, particularly the duration, which represents how long the person stayed at $l_j$. All time and locations adopt discretized representations where finer-grained ones increase computational complexity, while coarser ones may degrade performance. Each visit is represented by a simple text format -- \texttt{"arrival time is $t_j$, location is $l_j$, duration is $d_j$"}. The visits are separated by a special token and whitespaces, i.e., \texttt{"~=>~"}. Additionally, each text sequence representing a trajectory begins with \texttt{"<BOS>"} (Beginning Of Sentence token) and ends with \texttt{"<EOS>"} (End Of Sentence token). In this representation, the text sequence length depends only on the number of visits. Therefore, this method can effectively handle sporadic data. Meanwhile, previous methods heavily rely on fixed-interval location sequences, which can be challenging to manage high-resolution intervals or infrequent visits \cite{feng2018deepmove,yu2017seqgan}.

\partitle{Temporal-Order Permutation}. A critical challenge of decoder-only LLMs is their heavy dependence on the order of sequences. 
 To enable controllable generation and enable the LLM to learn the spatiotemporal correlations regardless of the order of the visits, we randomly permute the visits of $\mathbb{S}^{(i)}$. The permuted text sequence of $\mathbb{S}^{(i)}$ is denoted as $\mathbb{S}'^{(i)} = permute(\mathbb{S}^{(i)})$. This permutation only changes the order of visits without modifying any attributes or the order of tokens within each visit. Therefore, the semantic meaning remains the same. This way, the permuted data enhances the LLM's ability to learn the intrinsic properties and spatial-temporal relationships using the arrival times and durations rather than the visit order. For example, in the trajectory \texttt{"9am at work, 5pm at home"}, the specific order of visits is not inherently crucial to their semantic meaning. By permuting, the model can focus on directly learning the relationships between 
  location and time.

\partitle{Parameter-Efficient Fine-Tuning}. While we demonstrate \methodname using Llama2\cite{touvron2023llama}, an 
open-source decoder-only LLM, our framework can finetune any LLM using the permuted textual data. \methodname employs the standard causal language modeling task for fine-tuning, known as next-token prediction. The loss function is the cross entropy across the model's vocabulary, presented in Equation~\ref{eq:loss}, where $p$ is the probability, $N$ is the number of trajectories, $\theta$ stands for the trainable parameters, and $x_j$ represents tokens in the permuted text sequence $\mathbb{S}'^{(i)}$.

\vspace{-8mm}
\begin{equation}
\small
    \mathcal{L}(\theta) = -\sum_{i=1}^N \sum_{x_{j} \in \mathbb{S}'^{(i)}} \log p\left( x_j | x_{j-1}, x_{j-2}, ..., x_{1}, \theta \right)
    \label{eq:loss}
\end{equation}
\vspace{-3mm}


For computing efficiency, we employ a Parameter-efficient fine-tuning (PEFT) technique, known as Low Rank Adaption (LoRA)\cite{hu2021lora}. Specifically, given a pretrained weight matrix $W_0 \in \mathbb{R}^{dim \times k} $, LoRA introduces two trainable matrices $A \in \mathbb{R}^{r \times k} $ and $B \in \mathbb{R}^{dim \times r}$. $r$ is the rank, which is very small compared to $dim$ and $k$. This method constrains the update of $W_0$ by its low-rank decomposition $W_0 + \Delta W = W_0 + B A$. A regular forward pass of input $x$, $h = W_0 x$, can be modified using LoRA as:
\vspace{-2mm}
{\small
\[ h = W_0 x + \Delta W x = W_0 x + B A x \]
}
\vspace{-10mm}

\subsection{Generation}
Figure~\ref{fig:sampling} illustrates the generation process of \methodname, supporting both controlled and uncontrolled prompts. The fine-tuned LLM generates complete sequences based on the input prompts, which are then temporally reordered to construct the final trajectories. A trajectory integrity check is performed to ensure coherence in the generated trajectories, preventing issues such as overlapping time intervals between consecutive visits.

\partitle{Uncontrolled Prompting}. As all the text sequences for fine tuning start with \texttt{"arrival time is"}, uncontrolled prompting uses this exact phrase as prompts. The entire trajectory is then generated by the LLM.

\partitle{Controlled Prompting}. For controlled generation, initial prompts are created based on the given visit constraints, which means all visit constraints are strictly enforced. 
As mentioned in Def.~\ref{def:constraint}, a constraint $c$ includes a time period from $t_{c1}$ to $t_{c2}$. To simplify, we randomly sample a specific timestamp within the constrained period. We then convert the set of constraints into prompt using the same textual encoding that converts a trajectory into textual representation.  Next, we feed the prompts into the fine-tuned LLM. The remainder of the trajectory is then generated by the LLM. 
Our temporal-order permutation strategy enables the LLM to generate visits with arrival time both before or after the time of the constraints. It is worth noting that without permutation, the LLM would be restricted to generating visits that occur only after the constrained period, which would be limited in real-world trajectory generation. 

\partitle{LLM Generation}. Let $x^{(k)} = (x_{1}, x_{2}, ..., x_{k})$ be the current prompt, consisting of tokens after the tokenization. The LLM generation is an iterative process as follows. First, the probability of next tokens $p^{(k+1)}$ are computed by:
\vspace{-2mm}
{\small \[ p^{(k+1)} = \text{softmax} \left( \dfrac{f(x^{(k)}; \theta)}{T} \right), \]
\vspace{-5mm}
}

where $T$ is a temperature parameter, controlling the randomness of the sampling process. Higher $T$ increases diversity by exploring less likely tokens, which may lead to more varied but less coherent trajectories, while lower $T$ reduces randomness, favoring more likely tokens and resulting in more consistent but potentially repetitive trajectories. We sample based on the given probability $p^{(k+1)}$ to obtain the next predicted token $x_{k+1}$. The process then repeats with the updated prompt $x^{(k+1)} = (x^{(k)}, x_{k+1})$. This iterative procedure continues until the predicted token is \texttt{<EOS>}. Notably, due to the inherent randomness in sampling from $p^{(k+1)}$, the generated trajectories may vary even when the same prompts are used.

\partitle{Integrity Check.}
We first perform prompt format validation and reordering, followed by an integrity check to ensure the validity of the generated trajectories. Specifically, we detect and resolve overlaps in adjacent visit times, and correct out-of-bound location grid IDs by replacing them with nearby valid locations.

\vspace{-2mm}











%% file: 4.experiment.tex
\section{Experiments}


\indent In this section, we evaluate the performance of Geo-Llama on both real-world and synthetic datasets. We compare Geo-Llama with state-of-the-art mobility generation methods across uncontrolled, controlled, and data-efficiency study settings. Notably, our framework is uniquely designed to satisfy visit constraints, whereas most baseline methods, except Geo-CETRA, are not inherently suited for controlled generation. Since Geo-CETRA can only handle a single visit constraint during controlled generation, we include a stand-alone comparison with this baseline under the single-visit constraint setting. To ensure comparability in the controlled setting, we employ a straightforward approach by forcibly inserting (FI) the constrained visits into the trajectories generated by the baseline methods. Our code can be accessed at: 
\texttt{\url{https://github.com/Emory-AIMS/Geo-Llama}}.

    \begin{table}[ht]
    \centering
    \caption{Training Dataset Statistics for Geolife and MobilitySyn}
    \label{tab:dataset_statistics}
    \begin{tabular}{l|c|c}
        \hline
        \textbf{Dataset} & \makecell{\textbf{Total Daily}\\ \textbf{Trajectories Used}} & \makecell{\textbf{Average Stay Points}\\ \textbf{per Trajectory}} \\ \hline
        Geolife          & 7,000                                                     & 4.2                                             \\ 
        MobilitySyn      & 7,000                                                     & 6.7                                             \\ \hline
    \end{tabular}
\end{table}

\vspace{-4mm}

\begin{table*}[htb]
    \centering
    \scalebox{0.9}{
    \begin{tabular}{c|l|ccccc|cc}
       \multirow{2}{*}{\textbf{Dataset}} & \multirow{2}{*}{\textbf{Model}} & \multicolumn{5}{|c|}{\textbf{Trajectory-level} ($\downarrow$)} & \multicolumn{2}{c}{\textbf{Global-level} ($\downarrow$)}\\
       & & \textbf{Distance} & \textbf{G-radius} & \textbf{Duration} & \textbf{DailyLoc} & \textbf{I-rank} & \textbf{G-rank} & \textbf{Transition}\\ \hline
       \multirow{8}{*}{\textit{GeoLife}}
                & GRU & 0.0074 & 0.0630 & 0.0582 & 0.0839 & 0.0182 & 0.0139 & 0.0444 \\
                & LSTM & 0.0118 & 0.0594 & 0.1186 & 0.1197 & \textbf{0.0091} & 0.0139 & 0.0501 \\
                & Transformer & 0.0099 & 0.0460 & 0.1234 & 0.1413 & 0.0175 & 0.0173 & 0.0618 \\
                & VAE & 0.4350 & 0.6785 & 0.2660 & 0.6900 & 0.0414 & 0.1930 & 0.0647 \\
                & MoveSim & 0.0039 & 0.0447 & 0.0693 & 0.1875 & 0.0216 & 0.0207 & 0.0539\\
                & SeqGAN & 0.0076 & 0.1324 & 0.0541 & 0.3849 & 0.0273 & 0.0208 & 0.0599\\
                & Geo-CETRA & 0.0100 &    0.2479 & 0.1743 & 0.1168 &  0.0220 & \textbf{0.0000} & \textbf{0.0041} \\
                & Geo-Llama & \textbf{0.0010} & \textbf{0.0120} & \textbf{0.0014} & \textbf{0.0009} & 0.0161 & 0.0173 & 0.0267 \\ \hline
       \multirow{8}{*}{\textit{MobilitySyn}} 
                & GRU & 0.0093 & 0.1703 & 0.0617 & 0.1150 & \textbf{0.0000} & 0.0173 & 0.0085 \\       
                & LSTM & 0.0138 & 0.1716 & 0.1084 & 0.1842 & \textbf{0.0000} & 0.0182 & 0.0071 \\
                & Transformer &  0.0337 & 0.2642 & 0.1608 & 0.3590 & \textbf{0.0000} & 0.0236 & \textbf{0.0060} \\
                & VAE & 0.5347 & 0.6819 & 0.2817 & 0.6931 & 0.0249 & 0.2505 & 0.0127 \\
                & MoveSim & 0.0089 & \textbf{0.0474} & 0.5191 & 0.0948 & 0.0473 & 0.0207 & 0.0275\\
                & SeqGAN & 0.0063 & 0.1584 & 0.4637 & 0.0820 & 0.0034 & 0.0277 & 0.0109\\
                & Geo-CETRA & 0.0146 & 0.3620 & 0.2545 & 0.1706 & 0.0321 & \textbf{0.0000} & 0.0062\\
                & Geo-Llama & \textbf{0.0010} & 0.0689 & \textbf{0.0231} & \textbf{0.0181} & \textbf{0.0000} & 0.0139 & 0.0099 \\ \hline
    \end{tabular}}
    \caption{Results of uncontrolled trajectory generation.}
    \label{tab:uncond-exp}
\end{table*}

\subsection{Datasets \& Preprocessing}

\indent We conduct extensive experiments on the benchmark GeoLife dataset and a synthetic MobilitySyn dataset.

\begin{itemize}
    \item \textbf{GeoLife}. This GPS trajectory dataset \cite{zheng2010geolife} was collected by MSRA with 182 users over five years (from April 2007 to August 2012). It contains 17,621 trajectories, where each trajectory is a sequence of GPS records including the timestamp, latitude, and longitude. Here we only focus on data in the greater Beijing area.

    \item \textbf{MobilitySyn}. A realistic simulation of 5,000 agents performing daily movements across a metropolitan area over the course of a week. The simulation generates second-by-second GPS records for each agent, describing their recurring daily visits to various locations. The generated trajectories reflect plausible daily movement patterns.
\end{itemize}

\noindent \textbf{Staypoints Calculation.} Staypoints are defined as locations where an individual stays within a specified radius for a duration exceeding a predefined threshold. From the raw trajectory data, we extract staypoints from GPS records using a 1-kilometer radius for Geolife and a 0.1-kilometer radius for MobilitySyn, with a minimum duration of 15 minutes. The smaller radius for MobilitySyn accounts for its dense but geographically compact urban area, ensuring that small movements between nearby staypoints are not overlooked. In contrast, Geolife spans larger metropolitan regions, where a 1-kilometer radius better captures significant location transitions. The 15-minute threshold is used  across both datasets to reflect similar visit patterns. Next, we construct single-day trajectories, discarding those with fewer than three stay points. 
 Table \ref{tab:dataset_statistics} shows the number of daily trajectories we used and average number of staypoints per  trajectory in both datasets.
\\
\noindent \textbf{Trajectory Representations.} 
 Our baselines include approaches with both 1) fixed interval representation which are sequences of 96 visited locations per day, with fixed-length 15-minute intervals, and 2) sequence of visits representations. Geo-CETRA and Geo-Llama use sequences of visits, where the sequence length varies based on the number of visits, For Geo-Llama, each visit is represented by discretized location and time. The grid size is set to be a 1-kilometer box for Geolife and a 0.1-kilometer box for MobilitySyn, and the time interval is set to be 15-minutes. These setups align with the parameters used for staypoint calculation. Geo-CETRA uses continuous time and location, we perform the same discretization on its output for comparison. 
 
\partitle {Visit Constraint Generation.} For each stay point sequence, vist constraints are generated by randomly sampling a subset of visits within the trajectory. For each sampled visit, a corresponding location and time window are created, narrowly centered around the visit. This approach ensures that the visit constraints align with the actual data distribution and reflect realistic human mobility patterns.


\vspace{-3mm}

\subsection{Baselines}
\indent We evaluate the performance of our model against the following seven state-of-the-art baselines:
\begin{itemize}
    \item  \textbf{GRU}~\cite{lecun1998gradient} and \textbf{LSTM}~\cite{hochreiter1997long}: Recurrent neural networks that are efficient for sequential data generation. These models are able to predict the next location based on historically visited locations.
    \item \textbf{Transformer} \cite{vaswani2017attention}: A powerful deep learning model used in various natural language processing (NLP) and computer vision tasks that leverages self-attention mechanisms. A multi-layer Transformer decoder is utilized for location prediction (generation).
    \item  \textbf{Variational Autoencoders (VAE)} \cite{huang2019variational}: A generative model that learns to encode input data into a probabilistic latent space and reconstruct it. It converts the trajectories into 2D matrices, where each cell represents the location at a specific time step.
    \item  \textbf{SeqGAN} \cite{yu2017seqgan}: A sequence GAN that introduces a discriminator as a reward signal to guide the gradient policy update of the generator, which performs the next location prediction task based on the past states.

    \item  \textbf{MoveSim} \cite{feng2020learning}: An extension from SeqGAN. It introduces the attention module as the generator and incorporates domain knowledge such as  POI information in the model. For a fair comparison, we remove POI embedding in the MoveSim implementation, as we do not have access to POI information. 
    \item \textbf{Geo-CETRA} \cite{lin2024controllable}: A spatiotemporal point process-based framework for trajectory generation that incorporates constraint factorization and beam decoding to produce realistic trajectories under visit constraints. Due to the fact that it is limited to handling a single constraint and works on continuous geo-coordinates, we compare it with \methodname under the single visit constraint setting by aligning generated continuous coordinate with discrete grid IDs.
\end{itemize}

\begin{table*}[h]
    \centering
    \scalebox{0.9}{
    \begin{tabular}{c|l|ccccc|ccc}
       \multirow{2}{*}{\textbf{Dataset}} & \multirow{2}{*}{\textbf{Model}} & \multicolumn{5}{c|}{\textbf{Trajectory-level} ($\downarrow$)} & \multicolumn{3}{c}{\textbf{Global-level} ($\downarrow$)} \\ 
       & & \textbf{Distance} & \textbf{G-radius} & \textbf{Duration} & \textbf{DailyLoc} & \textbf{I-rank} & \textbf{G-rank} & \textbf{Transition} & \textbf{Top-K Transition} \\ \hline
       \multirow{7}{*}{\textit{\makecell{\textit{GeoLife} \\ (w 1-3 Visit \\ Constraints)}}}  
               & GRU {\footnotesize w/ FI} & 0.0078 & 0.0997 & 0.0479 & 0.1618 & 0.0157 & 0.0173 & 0.0385 & 0.0477 \\
                & LSTM {\footnotesize w/ FI} & 0.0145 & 0.1063 & 0.1151 & 0.2183 & 0.0115 & 0.0242 & 0.0438 & 0.0521\\
                & Transformer {\footnotesize w/ FI} & 0.0080 & 0.0458 & 0.0962 & 0.0980 & 0.0232 & 0.0242 & 0.0469 & 0.0555\\
                & VAE {\footnotesize w/ FI} & 0.3286 & 0.6452 & 0.2719 & 0.6849 & 0.0221 & 0.1324 & 0.0595 & 0.0781\\
                & MoveSim {\footnotesize w/ FI} & 0.0030 & 0.0173 & 0.0317 & 0.1646 & 0.0216 & 0.0139 & 0.0349 & 0.0456\\
                & SeqGAN {\footnotesize w/ FI} & 0.0051 & 0.0613 & 0.0140 & 0.3059 & 0.0364 & 0.0208 & 0.0333 & 0.0438 \\
                & Geo-Llama & \textbf{0.0007} & \textbf{0.0098} & \textbf{0.0014} & \textbf{0.0016} & \textbf{0.0061} & \textbf{0.0000} & \textbf{0.0209} & \textbf{0.0384} \\ \hline
       \multirow{7}{*}{\textit{\makecell{\textit{MobilitySyn} \\ (w 2-5 Visit \\ Constraints)}}}  
               & GRU {\footnotesize w/ FI} & 0.0089 & 0.2551 & 0.0381 & 0.1784 & \textbf{0.0000} & 0.0251 & 0.0069 & 0.0146 \\
                & LSTM {\footnotesize w/ FI} & 0.0132 & 0.2639 & 0.0806 & 0.2254 & \textbf{0.0000} & 0.0277 & 0.0063 & \textbf{0.0140}\\
                & Transformer {\footnotesize w/ FI} & 0.0243 & 0.2894 & 0.1225 & 0.3132 & \textbf{0.0000} & 0.0286 & \textbf{0.0057} & 0.0152\\
                & VAE {\footnotesize w/ FI} & 0.3044 & 0.6555 & 0.2624 & 0.6920 & 0.0248 & 0.0991 & 0.0124 & 0.0150 \\
                & MoveSim {\footnotesize w/ FI} & 0.0066 & 0.0815 & 0.0995 & 0.1000 & 0.0360 & 0.0243 & 0.0166 & 0.0464\\
                & SeqGAN {\footnotesize w/ FI} & 0.0051 & 0.1695 & 0.0487 & \textbf{0.0816} & 0.0034 & \textbf{0.0242} & 0.0076 & 0.0174 \\ 
                & Geo-Llama & \textbf{0.0021} & \textbf{0.0704} & \textbf{0.0217} & 0.1784 & \textbf{0.0000} & 0.0278 & 0.0073 & 0.0154 \\ \hline
    \end{tabular}}

    \caption{Results of controlled trajectory generation with multiple visit constraints. All baseline approaches forcibly insert (FI) specified visits from the constraints.}

    \label{tab:cond-exp}
\end{table*}

\begin{table*}[h]
    \centering
    \scalebox{0.9}{
    \begin{tabular}{c|l|ccccc|ccc}
       \multirow{2}{*}{\textbf{Dataset}} & \multirow{2}{*}{\textbf{Model}} & \multicolumn{5}{c|}{\textbf{Trajectory-level} ($\downarrow$)} & \multicolumn{3}{c}{\textbf{Global-level} ($\downarrow$)} \\ 
       & & \textbf{Distance} & \textbf{G-radius} & \textbf{Duration} & \textbf{DailyLoc} & \textbf{I-rank} & \textbf{G-rank} & \textbf{Transition} & \textbf{Top-K Transition} \\ \hline
       \multirow{2}{*}{\textit{GeoLife}}  
               & Geo-CETRA {\footnotesize w/ Single Constraint} & 0.0093 & 0.2772 & 0.1789 & 0.1091 & \textbf{0.0140} & \textbf{0.0005} & \textbf{0.0038} & \textbf{0.0324} \\
                & Geo-Llama {\footnotesize w/ Single Constraint} & \textbf{0.0011} & \textbf{0.0132} & \textbf{0.0011} & \textbf{0.0028} & 0.0174 & 0.0208 & 0.0295 & 0.0581 \\ \hline
       \multirow{2}{*}{\textit{MobilitySyn}}  
                & Geo-CETRA {\footnotesize w/ Single Constraint}  & 0.0139 & 0.4054 & 0.2616 & 0.1601 & 0.0208 & \textbf{0.0000} & \textbf{0.0059} & 0.1036\\ 
                & Geo-Llama {\footnotesize w/ Single Constraint} & \textbf{0.0024} & \textbf{0.0902} & \textbf{0.0246} & \textbf{0.2038} & \textbf{0.0000} & 0.0278 & 0.0083 & \textbf{0.0188} \\ \hline
    \end{tabular}}

    \caption{Results of controlled trajectory generation under single visit constraint. Geo-CETRA is designed to handle only one constraint.}
    
    \label{tab:cond-exp-geo-cetra}
    \vspace{-6mm}
\end{table*}

\vspace{-1mm}
\subsection{Evaluation Metrics}
As constraints are strictly enforced and satisfied in our controllable generation framework, we focus our evaluation on the realism and contextual coherence of the generated trajectories rather than constraint satisfaction. Following previous works \cite{feng2020learning,zhang2023csgan}, we employ a variety of metrics to assess the quality of the generated trajectories by comparing key mobility pattern distributions between the generated and real trajectories. 







\textbf{Distance} is the distribution of cumulative travel distance per user each day. \textbf{G-radius} (radius of gyration) represents the distribution of spatial range of user's daily movement. \textbf{Duration}  is the distribution of the duration per visited location. \textbf{DailyLoc}  is the distribution of the number of visited locations per day for each user. \textbf{G-rank} is the global distribution of number of visits per location for top-100 visited locations. \textbf{I-rank} is an individual version of the G-rank.


For the above properties, the Jensen-Shannon divergence, 
$\text{JSD}\left(\mathbb{D}, \mathbb{D}'\right) = h\left(\frac{\mathbb{D} + \mathbb{D}'}{2}\right) - \frac{h(\mathbb{D}) + h(\mathbb{D}')}{2}$, is applied to measure the distance or discrepancy between the distributions of generated $\mathbb{D}'$ and real trajectory $\mathbb{D}$, where \( h \) represents the Shannon information.  The lower the JSD, the more realistic the generated trajectories. 


In addition, we also have metrics for comparing transition distribution between the generated and real trajectories. \textbf{Transition} is the probability distribution of a trajectory transitioning from location $l_1$ to location $l_2$ over the set of all discretized locations $\mathcal{G}$. \textbf{Top-K Transition} are transitions involving top-K locations included in the constraints, which we introduce to better evaluate whether the methods incorporate the constraints with realistic transitions.



We use the Frobenius norm of their difference: $\|P_\mathbb{D} - P_\mathbb{D'}\|_F = \sqrt{\sum_{l_1=1}^{|\mathcal{G}|} \sum_{l_2=1}^{|\mathcal{G}|} \left| P_\mathbb{D}(l_1, l_2) - P_\mathbb{D'}(l_1, l_2) \right|^2}$
as the discrepancy metric. The lower the better. We note that G-rank and Transition correspond to global level while others correspond to trajectory-level patterns.

\par\vspace{-3mm}
\subsection{Implementation Details}

\begin{itemize}
    \item \textbf{GRU, LSTM, and Transformer} 
are trained for 200 epochs using the Adam optimizer with a learning rate of 0.001. Models share an embedding size of 256, with GRU and LSTM using 6 layers of 512 hidden units, while Transformer adopts a decoder-only architecture with 4 layers and 4 attention heads.

    \item \textbf{VAE} 
    uses three fully connected layers in both the encoder and decoder, each with 128 units. The latent dimension is 64, and training was conducted over 200 epochs with a batch size of 256.

    \item \textbf{SeqGAN}
includes an LSTM-based generator trained with 16-dimensional embeddings and 16 hidden units. Discriminator employs diverse filter sizes and counts, with 40 epochs of pre-training and 20 epochs of adversarial training.

    \item \textbf{MoveSim} 
    builds upon SeqGAN with similar parameter settings but adds an attention-based generator with a location embedding size of 256, a time embedding size of 16, and a hidden dimension of 64.

    \item \textbf{Geo-CERTA}
    employs a conditional Gaussian Mixture Model  with 8 spatial and temporal mixture components and a beam search strategy with a beam size 10 and top $k$=3. 
    Optimization is performed with the Adam optimizer, a learning rate of 0.01, a scheduler with a decay factor of 0.99, and z-score normalization for input data.
    
    \item \textbf{\methodname} 
    fine-tunes the Llama-2-7b-chat-hf model using LoRA with a batch size of 48, a learning rate of 0.00001, LoRA alpha32, LoRA dropout 0.02, LoRA r 16, and 20 epochs. Sampling uses a temperature of 1.2 by default. We note that our framework is general and can use any pretrained LLMs. 
\end{itemize}

\subsection{Results}

\partitle{Uncontrolled Generation}. Table~\ref{tab:uncond-exp} compares \methodname with all baselines under the uncontrolled setting. It shows that \methodname consistently outperforms all baselines across the majority of location-based metrics (Distance, G-radius, G-rank, and I-rank), time-based metrics (Duration and DailyLoc), as well as transition metrics. This highlights the superior capability of the finetuned LLM of \methodname in capturing the spatiotemporal patterns compared to other models. The significant performance advantage on time-based metrics can be also attributed to Geo-Llama's representation of the trajectory as a sequence of visit with time and location as separate features (in contrast to the fixed interval representation), resulting in shorter sequences, averaging 5 visits per trajectory. This advantage can be also observed in Geo-CETRA which also uses sequence of visit representation. 
In contrast, most other baseline methods use 96-time-step fixed intervals, which may obscure temporal patterns due to the repeated locations for a single staypoint, although some still perform adequately on location frequency metrics. 

Among the baselines, VAE struggled to converge and effectively model mobility patterns potentially because sequences are transformed into an ill-suited 2D image format. Despite relying on the sequence model as the generator, SeqGAN and MoveSim encountered the inherent training instability of GAN-based methods, which yields poor sampling diversity and degraded performance on both time and location metrics. On the other hand, GRU, LSTM, and Transformer not only effectively capture location-based patterns, but also outperform VAE, MoveSim, and SeqGAN on time-based metrics, underscoring their  effectiveness in handling temporal information. 

{\textit{In uncontrolled settings, we conclude that \methodname consistently outperforms the baselines in generating realistic trajectories across a range of metrics and datasets}}.

\partitle{Controlled Generation}. We compare \methodname with all baselines under the controlled setting.  We separate the comparison under multiple and single visit constraints. For multiple visit constraints, we use 1-3 visit constraints per trajectory for GeoLife dataset and 2-5 visit constraints for MobilitySyn dataset respectively. We will also evaluate the impact of number of visit constraints in later experiments.  All baselines forcibly insert (FI) the specified visits after generation to satisfy the constraints. For single constraint, we compare with Geo-CETRA since it only supports a single visit constraint. 

\subsubsection{Comparison of Geo-Llama under Multiple Visit Constraints}

Table~\ref{tab:cond-exp} demonstrates that the trajectories generated by Geo-Llama under multiple visit constraints per trajectory remain contextually coherent, and reflect true underlying spatiotemporal patterns. In contrast, the trajectories generated by baseline models, display unrealistic time, and transition patterns due to the forcible insert of constrained visits. Geo-Llama’s ability to model spatiotemporal dynamics independently of sequence order, relying solely on the temporal attributes of each event, allows it to integrate external visit constraints in a context-aware manner, enabling the generation of realistic trajectories. 
In addition, while forcibly inserting specified visits from constraints can boost location-based metrics (e.g., G-rank and I-rank) of baseline methods by introducing commonly occurring real-world locations, these approaches fail to preserve correct temporal order and smooth transitions between visits because the forcible insertion disrupts the realistic spatiotemporal relationships, leading to unrealistic and contextually incoherent trajectories. 

{\textit{Overall, by leveraging our temporal-order-permutation strategy, \methodname not only consistently enforces the specified constraints but also generates more realistic trajectories than baselines}}.

\begin{figure*}[htp]
\centering
\begin{subfigure}{.23\textwidth}
  \centering
  \includegraphics[width=\linewidth]{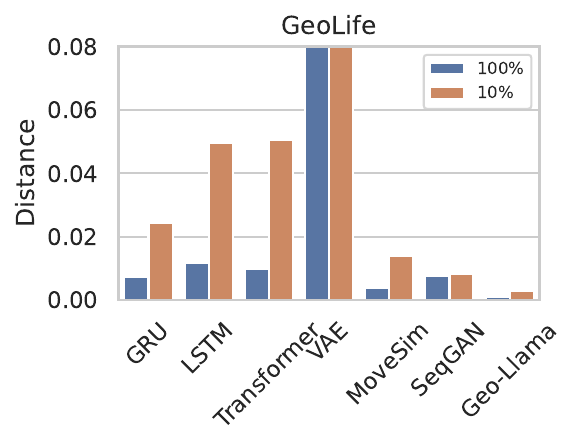}
\end{subfigure}%
\begin{subfigure}{.23\textwidth}
  \centering
  \includegraphics[width=\linewidth]{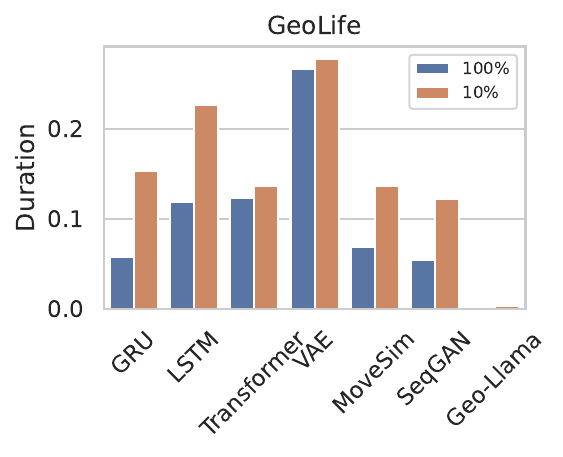}
\end{subfigure}
\begin{subfigure}{.23\textwidth}
  \centering
  \includegraphics[width=\linewidth]{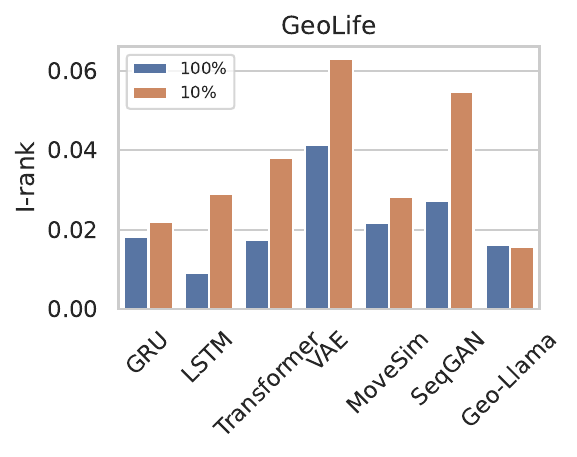}
\end{subfigure} 
\begin{subfigure}{.23\textwidth}
  \centering
  \includegraphics[width=\linewidth]{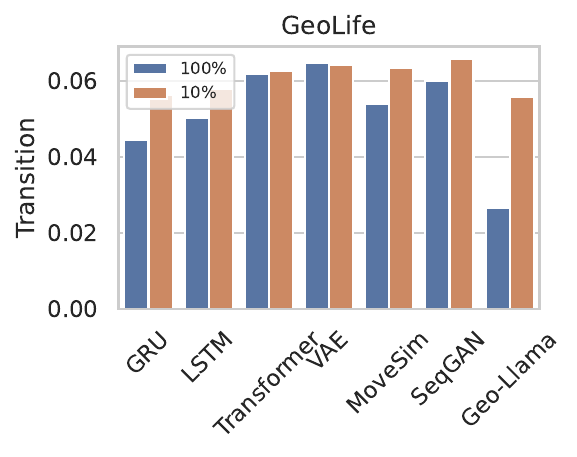}
\end{subfigure} \\
\begin{subfigure}{.23\textwidth}
  \centering
  \includegraphics[width=\linewidth]{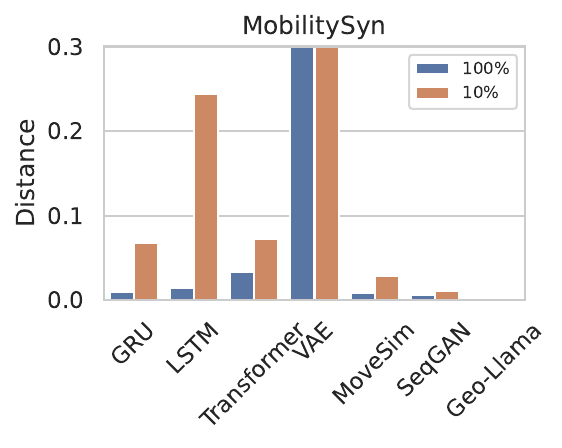}
\end{subfigure}
\begin{subfigure}{.23\textwidth}
  \centering
  \includegraphics[width=\linewidth]{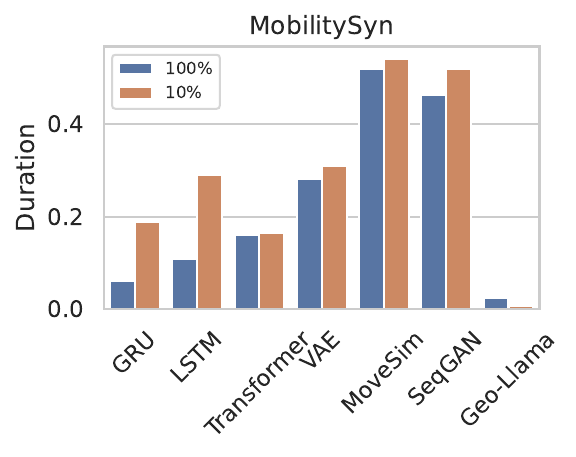}
\end{subfigure}
\begin{subfigure}{.23\textwidth}
  \centering
  \includegraphics[width=\linewidth]{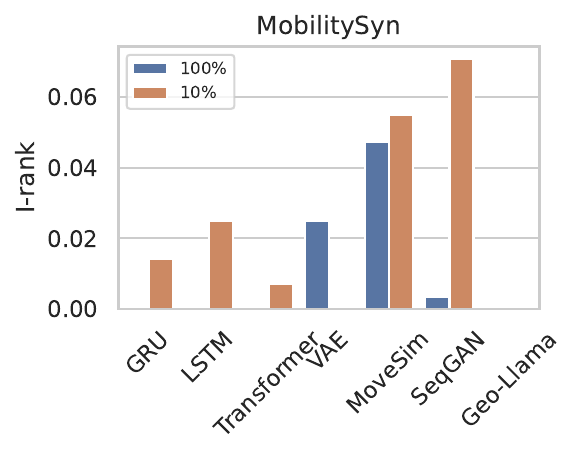}
\end{subfigure}
\begin{subfigure}{.23\textwidth}
  \centering
  \includegraphics[width=\linewidth]{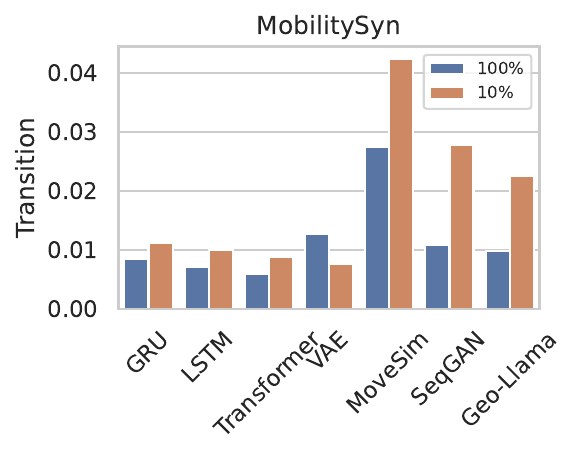}
\end{subfigure}%
\caption{Data-efficient learning study. We crop some extremely high values of VAE for better visualization.}
\label{fig:data-efficient}
\vspace{-3mm}
\end{figure*}

\subsubsection{Comparison of Geo-Llama under Single Visit Constraint with Geo-CETRA}

Table \ref{tab:cond-exp-geo-cetra} indicates that Geo-Llama demonstrates superior performance in both spatial (Distance, G-radius) and temporal (Duration, DailyLoc) metrics compared to Geo-CETRA, indicating more consistent modeling of spatiotemporal patterns. While Geo-CETRA excels in transition-based metrics, particularly Top-K Transition, due to its constraint decomposition algorithm's ability to capture realistic transitions between constrained and subsequent events, this comes at the cost of increased spatial and temporal distortion. The higher G-radius and lower DailyLoc for Geo-CETRA highlight its inability to maintain precise spatial locations and daily visit patterns. Furthermore, Geo-CETRA's constraint decomposition algorithm is inherently inflexible, supporting only a single visit constraint, which limits its applicability for more complex trajectory generation tasks.
\textit{Overall, Geo-Llama outperforms Geo-CETRA in producing more context-coherent trajectory under a single visit constraint and is also more flexible in taking more complicated visit constraints.}

\partitle{Data-efficient Learning}.
We train \methodname and all baselines on 10\% of both datasets ($\sim$700 trajectories) to evaluate their data-efficient learning capabilities. Figure~\ref{fig:data-efficient} presents the model performance in different metrics using 100\% and 10\% of the data. 
When trained on 10\% of the dataset, \methodname has only a slight performance drop across most metrics compared to training on the full dataset. In contrast, baseline models have a significant performance drop as the training data size decreases. For instance, the distance JSD of the baseline methods increases by 2 to 10 times, while \methodname's remains stable. This data-efficient feature of \methodname may result from leveraging pretrained LLMs, whereas baseline models require training from scratch. The results demonstrate that {\textit{\methodname is data-efficient whereas the baseline methods show significant performance degradation with limited training data}}.

\vspace{-1mm}
\section{Parameter Studies}

\begin{figure}[htp]
   \vspace{-4mm}
    \centering
    \begin{subfigure}{.24\textwidth}
        \centering
        \includegraphics[width=\linewidth]{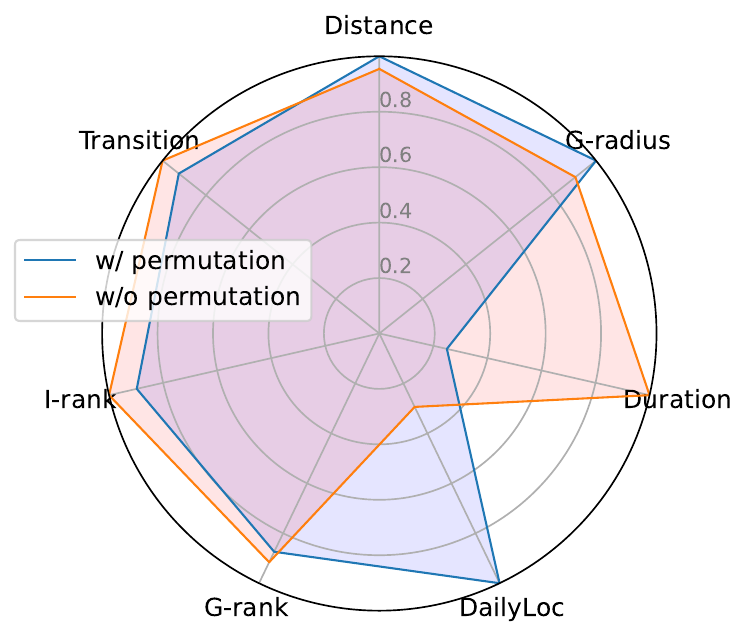}
        \caption{Uncontrolled generation.}
        \label{fig:permute-no-permute}
    \end{subfigure}
    \begin{subfigure}{.24\textwidth}
        \centering
        \includegraphics[width=\linewidth]{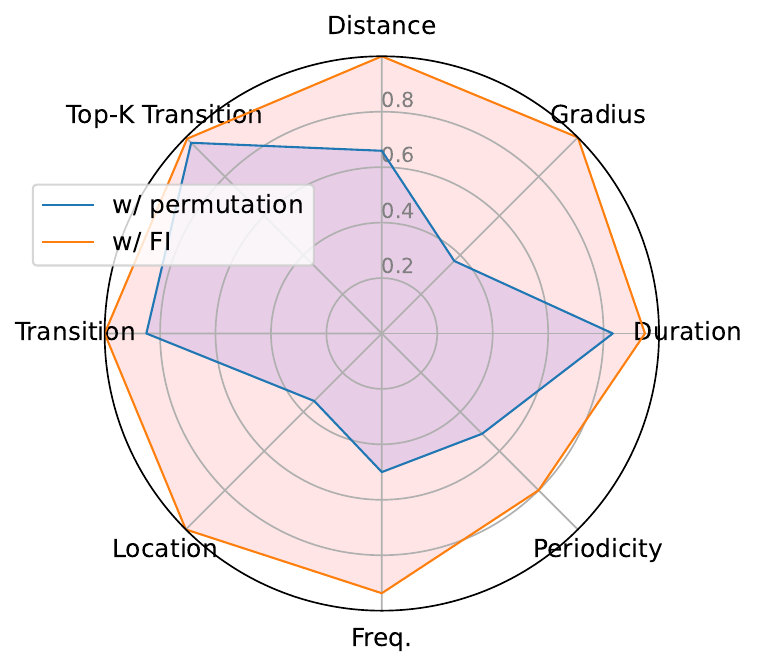}
        \caption{Controlled generation.}
        \label{fig:permute-RI}
        \end{subfigure}
    \caption{Impact of temporal-order permutation. The reported performance metrics represent average values across both datasets. For each dataset, the values are scaled to a range of 0 to 1 by dividing them by their respective maximum values.}
    \label{fig:permutation}
\end{figure}

\partitle{Impact of Temporal-Order Permutation}. To validate our hypothesis that adopting temporal-order permutation enhances the capture of true temporal features among visits regardless of sequence orders, we train two identical models on the same dataset and with same hyperparameter setting: one incorporating permutation and the other not. Figure~\ref{fig:permute-no-permute} shows the results for uncontrolled generation. Permutation has a negligible effect on most metrics, including G-radius, Distance, G-rank, G-rank, and Transition. However, notable differences are observed in Duration and DailyLoc. With permutation, \methodname performs better on Duration, likely because it allows the model to capture the relationship between arrival time and duration without relying on sequence order. In contrast, the model without permutation outperforms on DailyLoc, as preserving temporal order simplifies the recognition of periodic patterns.\\

\vspace{-2mm}
For controlled generation without temporal-order permutation, even if we can still use the constraints as prompts, \methodname would not generate realistic trajectories because all the generated visits will be after the arrival time of the last constrained visit. To construct a baseline, we first conduct uncontrolled generation using \methodname without permutation and then forcibly insert (FI) the constrained visits into the trajectories. Figure~\ref{fig:permute-RI} illustrates each method's performance, demonstrating that \methodname with permutation significantly outperforms the forcible-insertion approach across all metrics. Notably, employing permutation yields around 80\% improvement in both G-rank and G-radius. Overall, for uncontrolled generation, \textit{\methodname yields competitive performance with or without permutation. However, for controlled generation, the proposed permutation strategy enables LLMs to generate more realistic trajectories compared to forcible insertion}.





\begin{figure}[htp]
    \centering
   \vspace{-4mm}
    \begin{subfigure}{.24\textwidth}
        \centering
        \includegraphics[width=\linewidth]{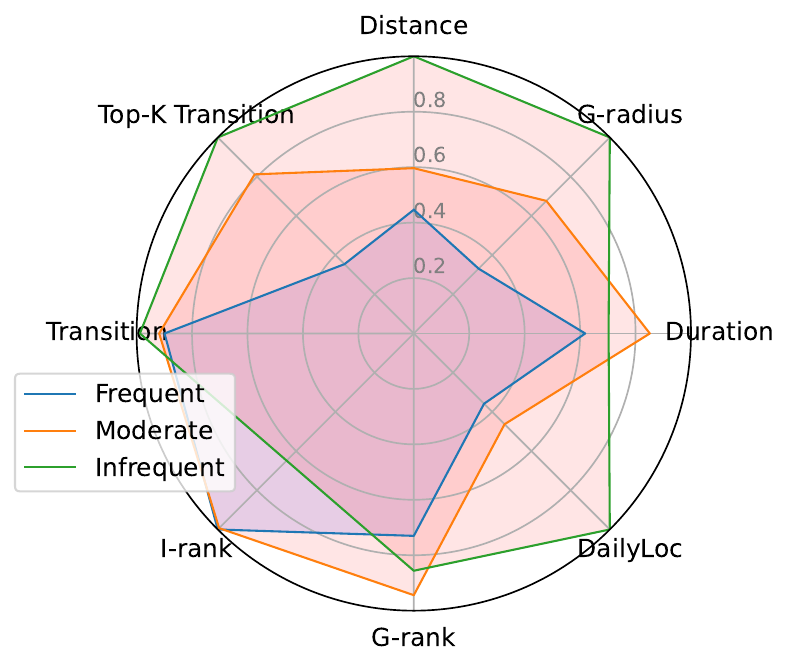}
        \caption{Varying location frequency.}
        \label{fig:location-freq}
    \end{subfigure}
    \begin{subfigure}{.24\textwidth}
        \centering
        \includegraphics[width=\linewidth]{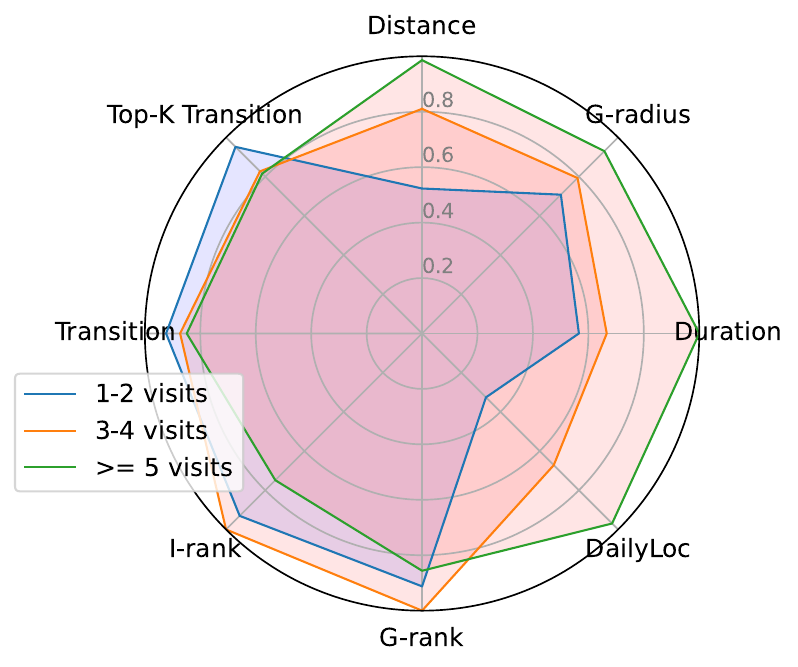}
        \caption{Varying numbers of visits.}
        \label{fig:num-visit}
    \end{subfigure}
    \caption{Impact of constraint visits. The reported performance metrics represent average values across both datasets, with each metric scaled by its maximum value within each dataset.}
   \vspace{-5mm}    
\end{figure}

\partitle {Impact of Location Popularity in Constraints}. Intuitively, constraints involving popular (frequently visited) locations are easier to satisfy. To investigate this relationship, we categorize constraints into three types based on location visit counts in the training data: frequent, moderate, and infrequent. We then rank locations by their total visits and select the top 40 (most visited), middle 40, and bottom 40 (least visited) to construct the corresponding constraint sets. Figure~\ref{fig:location-freq} shows that the synthetic trajectories are more realistic when their constraints contain more frequently visited locations, as expected. Notably, an 80\% performance disparity exists between frequent and infrequent constraints for Top-K Transition, G-radius, and DailyLoc, potentially because more frequent locations appear more in the training data, enabling the LLMs to learn patterns of those locations more effectively. Overall, \textit{\methodname tends to perform better with constraints involving more frequently visited locations}.

\partitle{Impact of Number of Visits in Constraints}. Figure~\ref{fig:num-visit} illustrates the performance of \methodname with varying number of visits in the constraints.  Trajectories containing 1-2 constrained visits exhibit significantly lower discrepancy in Distance, G-radius, Duration and DailyLoc --- about 40\% less than those with 3-4 visits and 20\% less than those with at least 5 constrained visits. Intuitively,  fewer constrained visits allow LLMs greater flexibility during generation. Therefore, \textit{fewer constrained visits generally yield more realistic generated trajectories}.

\begin{figure}[htp]
    \centering
    \begin{minipage}[b]{0.49\linewidth}
        \centering
        \includegraphics[width=\linewidth]{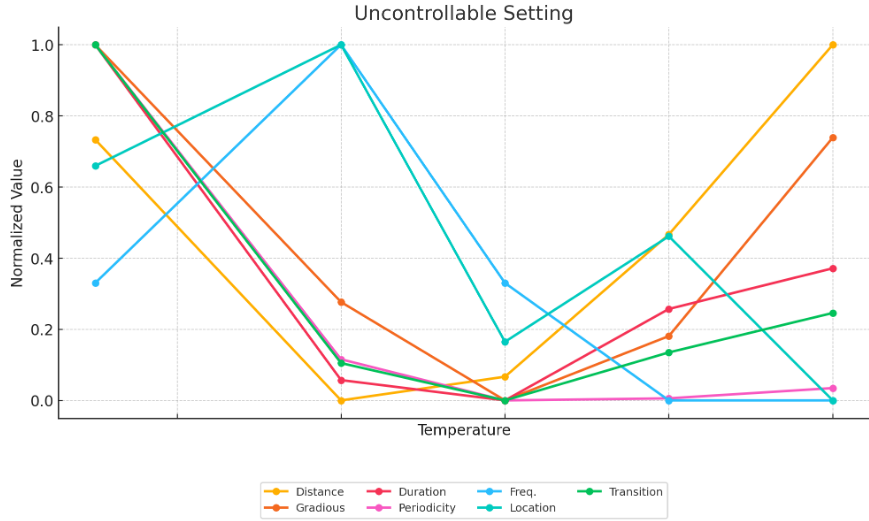}
        \vspace{-2mm}
        \caption*{\small (a) Uncontrollable Setting.}
    \end{minipage}
    \hfill
    \begin{minipage}[b]{0.49\linewidth}
        \centering
        \includegraphics[width=\linewidth]{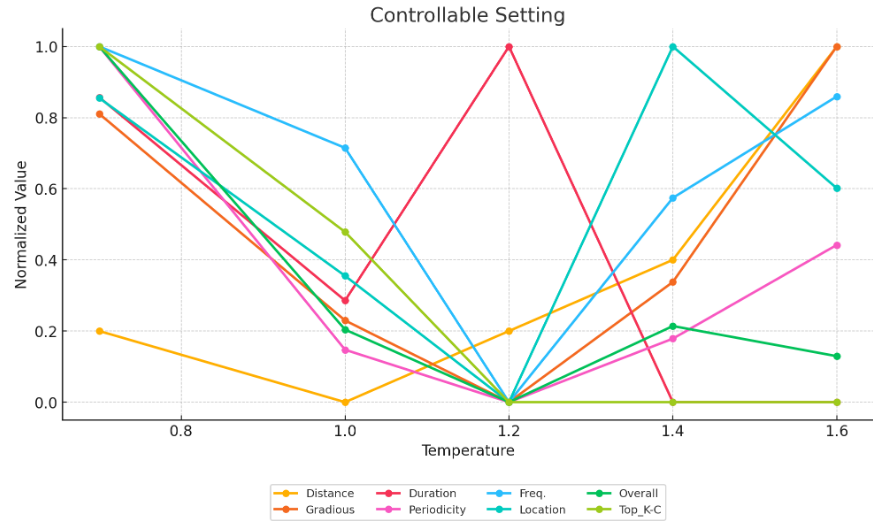}
        \vspace{-2mm}
        \caption*{\small (b) Controllable Setting.}
    \end{minipage}
    \vspace{-3mm}  
    \caption{Impact of temperature on generation. The normalized values are averaged across two datasets and scaled to the [0,1] range.}
    \label{fig:temp-combined}
    \vspace{-2mm} 
\end{figure}

\partitle{Impact of Temperature.} Temperature is an important hyper-parameter in LLM sequence generation, controlling randomness: higher values increase diversity but risk incoherence, while lower values favor consistency but may lead to repetition. Figures~\ref{fig:temp-combined} depicts the performance of \methodname with temperature increasing from 0.7 to 1.6 for both uncontrolled and controlled generation. Most metrics initially decrease to reach their lowest points around a temperature of 1.2 and then recover as the temperature continues to rise. This result indicates that \textit{temperature significantly affects  performance, the optimal temperature is approximately the same at 1.2 for  uncontrolled and controlled generation with a good tradeoff between diversity and consistency}.


%% file: 5.conclusion.tex
\vspace{-4mm}
\section{Conclusion}
We have presented Geo-Llama, a novel framework that employs LLM fine-tuning to address the challenges of realistic trajectory generation with spatiotemporal constraints such as enforcing specific visits. By utilizing a visit-wise permutation strategy for trajectory fine-tuning, our method ensures the generation of high-quality, contextually coherent synthetic trajectories that strictly adhere to the constraints. Our approach not only outperforms existing approaches in capturing complicated mobility patterns and generating realistic trajectories in a more data-efficient way, but also overcomes the limitations of current controllable sequence generation techniques, which can only handle semantic-level and implicit constraints or a single explicit constraint. 

In future work, we plan to extend Geo-Llama by incorporating auxiliary information, such as point of interest (POI) and modality features, and by utilizing continuous time and location representations. These enhancements aim to improve the modeling of more realistic and complex trajectories. In addition, we plan to explore domain transfer techniques to enable adaptation to unknown regions with limited data, investigate the impact of model size by comparing different LLMs, and develop prompt compression methods such as soft prompt tuning to reduce input length and computational cost.


%% file: acknowledgment.tex
\section*{Acknowledgments}
Research supported by the Intelligence Advanced Research
Projects Activity (IARPA) via the Department
of Interior/Interior Business Center (DOI/IBC) contract
number 140D0423C0033. The U.S. Government is authorized
to reproduce and distribute reprints for Governmental
purposes, notwithstanding any copyright annotation
thereon. Disclaimer: The views and conclusions contained
herein are those of the authors and should not
be interpreted as necessarily representing the official
policies or endorsements, either expressed or implied, of
IARPA or the U.S. Government.